\documentclass{article}

\usepackage[preprint]{neurips_2025}

\usepackage[utf8]{inputenc} 
\usepackage[T1]{fontenc}   
\usepackage{hyperref}       
\usepackage{url}            
\usepackage{booktabs}       
\usepackage{amsfonts}      
\usepackage{nicefrac}       
\usepackage{microtype}     
\usepackage{xcolor}         
\usepackage{graphicx}
\usepackage{enumitem} 
\usepackage{listings}
\usepackage{subcaption}
\usepackage{tikz}
\usepackage{float}

\title{Visual serial processing deficits explain divergences in human and VLM reasoning}

\author{%
  Nicholas Budny\textsuperscript{1\textsuperscript{*}}, Kia Ghods\textsuperscript{1\textsuperscript{*}}, Declan Campbell\textsuperscript{1\textsuperscript{*}}, Raja Marjieh\textsuperscript{2}, Amogh Joshi\textsuperscript{1}, Sreejan Kumar\textsuperscript{1}\\
  \textbf{Jonathan D. Cohen\textsuperscript{1,2\textsuperscript{\dag}}, Taylor W. Webb\textsuperscript{3,4\textsuperscript{\dag}}, and Thomas L. Griffiths\textsuperscript{2,5\textsuperscript{\dag}} } \\ \\
  \textsuperscript{1}Princeton Neuroscience Institute\\
  \textsuperscript{2}Department of Psychology, Princeton University\\
  \textsuperscript{3}Department of Psychology, Université de Montréal \\
  \textsuperscript{4}Mila - Quebec AI Institute \\
  \textsuperscript{5}Department of Computer Science, Princeton University\\
  \textsuperscript{*}Equal contribution\\
  \textsuperscript{\dag}Equal advising
  \\
}

\begin{document}

\maketitle

\begin{abstract}
Why do Vision Language Models (VLMs), despite success on standard benchmarks, often fail to match human performance on surprisingly simple visual reasoning tasks? While the underlying computational principles are still debated, we hypothesize that a crucial factor is a deficit in visually-grounded serial processing. To test this hypothesis, we compared human and VLM performance across tasks designed to vary serial processing demands in three distinct domains: geometric reasoning, perceptual enumeration, and mental rotation. Tasks within each domain varied serial processing load by manipulating factors such as geometric concept complexity, perceptual individuation load, and transformation difficulty. Across all domains, our results revealed a consistent pattern: decreased VLM accuracy was strongly correlated with increased human reaction time (used as a proxy for serial processing load). As tasks require more demanding serial processing---whether composing concepts, enumerating items, or performing mental transformations---the VLM-human performance gap widens reliably. These findings support our hypothesis, indicating that limitations in serial, visually grounded reasoning represent a fundamental bottleneck that distinguishes current VLMs from humans.
\end{abstract}

\section{Introduction}

Large-scale vision-language models (VLMs), such as GPT-4o \citep{achiam2023gpt}, Claude-Sonnet 3.7 \citep{anthropic2023claude}, and Gemini 2.5 Pro \citep{team2023gemini}, demonstrate impressive capabilities, often matching or surpassing human-level accuracy on complex benchmarks like image captioning and visual question answering \citep{fu2023mme}. Yet, paradoxically, they also exhibit surprising deficits in tasks that appear comparatively simple, such as counting and visual search \citep{campbell2025understandinglimitsvisionlanguage}. Here, we hypothesize that these discrepancies, and a large amount of VLM-human divergence more broadly, stem from deficits in visually-grounded serial processing. To investigate this hypothesis and better evaluate the extent of these deficits, we use tasks from cognitive science \citep{dehaene2006core} to vary serial processing demands systematically.% and measure the corresponding deviations in VLM performance relative to human baselines.

To provide a proxy for serial processing load, we use human reaction time (RT) --- a well-established behavioral marker of serial cognitive engagement. Humans adapt to complexity in visual reasoning by trading time for accuracy, using sequential attention to focus on different elements when analyzing complex scenes \citep{heitz2014speed, stewart2020review}. We hypothesize that VLMs largely lack this capability, as their reasoning is tethered to sequential generation of text (their chain of thought), not sequential analysis of the image. We therefore predict that VLM accuracy will be inversely correlated with human RT, indicating that as tasks require more extensive serial processing by humans, the VLM-human performance gap widens. This method allows us to identify where VLM performance diverges most significantly from human performance and to evaluate whether these divergences align with visual concepts and operations hypothesized to demand more intensive serial processing. 

To test our hypothesis, we systematically manipulate the serial processing load in three distinct  domains. First, in a geometric program induction task inspired by work modeling geometric concepts as programs \citep{hsu2022geoclidean}, we vary load by manipulating concept complexity, operationalized as the number of geometric primitives (Minimum Description Length; MDL) required to define a geometric concept \citep{SABLEMEYER2022101527}. Second, for visual enumeration, drawing on work distinguishing subitizing from serial counting \citep{trick1994small}, serial load is varied by manipulating color distinctiveness and spatial overlap, known to increase demands on serial attention \citep{franconeri2013flexible}. Finally, in a mental rotation task based on \citep{shepard1971mental}, serial processing demands are modulated by varying the angular disparity between two stimuli, a factor shown to linearly increase human processing time \citep{cooper1973chronometric}. Across these diverse tasks we demonstrate that as task conditions necessitate greater serial processing %--- whether composing concepts, enumerating items, or performing mental transformations --- 
the VLM-human performance gap widens reliably, providing convergent evidence for a serial processing deficit.

\section{Background and Related Work}

\subsection{Vision Language Models}

Vision Language Models (VLMs) are designed to jointly model visual and linguistic information, typically by learning multimodal representations that bridge inputs from both domains \citep{Radford2021CLIP, Alayarac2022Flamingo}. These models commonly incorporate a vision encoder to extract features from images or videos, which are then interfaced with a large language model conditioned on these visual features alongside textual prompts \citep{achiam2023gpt, team2023gemini, anil2023palm2, meta2024llama4}. This architectural paradigm has enabled significant progress on a variety of downstream tasks, including image captioning, visual question answering (VQA), and, more recently, complex multimodal reasoning \citep{Liu2023LLaVA, Zhu2023MiniGPT4}. Despite these advances, the precise mechanisms by which these models integrate and reason about visual information remain an active area of study. Of particular interest is the extent to which the reasoning strategies learned by VLMs correspond to, or diverge from, human visual reasoning strategies.

\subsection{Measuring correspondences between human visual processing and VLMs}
To gauge the alignment between human and VLM visual reasoning, recent studies have benchmarked VLMs against human performance on a wide range of visual reasoning tasks from cognitive science. These comparisons have uncovered both similarities and crucial differences in their behavioral biases \citep{campbell2025understandinglimitsvisionlanguage, dasgupta2022language}.
In some geometric reasoning tasks, VLMs exhibit remarkably human-like biases \citep{campbell2023relationalconstraintsneuralnetworks, campbell2024humanlikegeometricabstractionlarge}, successfully performing tasks involving perceptual grouping and identifying partially obscured objects \citep{lampinen2025inductive, pothiraj2025captureevaluatingspatialreasoning}. The relative difficulty of spatial reasoning tasks also aligns between VLMs and humans \citep{xu2025definingevaluatingvisuallanguage}, suggesting some commonalities.  

Despite these similarities, a significant gap persists between VLM and human performance across several core cognitive tasks \citep{SchulzeBuschoff2025}. VLMs are known to have deficits in spatially grounding reasoning \citep{tong2024cambrian1, kamath2023whats, xu2025definingevaluatingvisuallanguage, ramakrishnan2025does}, feature integration \citep{campbell2025understandinglimitsvisionlanguage}, and multi-step reasoning \citep{campbell2025understandinglimitsvisionlanguage, greff2020binding}. These manifest in specific tasks where VLM performance is near-chance: determining whether circles overlap, counting, identifying circled letters \citep{rahmanzadehgervi2025visionlanguagemodelsblind}, estimating depth, establishing visual correspondence, performing multi-view reasoning \citep{fu2024blinkmultimodallargelanguage}, and even failing at simple visual analogies that are solvable by toddlers \citep{goddard2025kiva}—all tasks where human accuracy is at ceiling.

What unites these seemingly diverse failures? This constellation of behavioral traits \citep{ramakrishnan2025does} currently lacks a unifying explanation. While various frameworks have catalogued VLM failures across different benchmarks \citep{SchulzeBuschoff2025, ramakrishnan2025does, huang2025visionlanguagemodelsstruggle}, effectively demonstrating \textit{when} performance falters, the question of \textit{why} they fail in these settings remains unresolved. Existing explanations range from training data constraints to deficits in core knowledge, but no single framework accounts for the breadth of observed failures. We propose that a deficit in serial processing provides this missing unifying framework.

\section{The serial processing deficit hypothesis}

We hypothesize that these diverse perceptual failures stem from a fundamental deficit in visually grounded serial processing. Three observations support this hypothesis: VLMs perform well when visual elements are well-separated but fail when elements require sequential analysis \citep{rahmanzadehgervi2025visionlanguagemodelsblind}; vision encoders contain sufficient information to solve these tasks, yet language models fail to effectively decode this information \citep{rahmanzadehgervi2025visionlanguagemodelsblind}; and tasks that cannot be easily decomposed into linguistic steps pose particular challenges \citep{fu2024blinkmultimodallargelanguage}.

Our contribution is threefold: We provide the first investigation of serial processing as a unifying explanation for VLM failures across diverse visual reasoning tasks. We demonstrate that human RT serves as a reliable predictor of VLM performance deficits, offering a way to predict model failures on novel tasks. And finally, we show that these limitations hold across different model architectures and prompting strategies, suggesting a fundamental limitation rather than a superficial training gap.

\subsection{Serial processing in VLMs}

Inductive biases encouraging serial, step-by-step processing have been instrumental in unlocking advances in the mathematical reasoning and coding capabilities of Large Language Models (LLMs). The Tree of Thoughts framework \citep{yao2023tree}, for instance, enables LLMs to systematically explore, evaluate, and prune multiple reasoning pathways, tackling problems that were previously intractable. The success of such methods in language, where decomposing complex tasks dramatically boosts performance, naturally motivated exploration of analogous strategies for Vision-Language Models (VLMs).
Initial attempts to instill serial processing in VLMs --- through fine-tuning \citep{deitke2024molmo} and tool use \citep{yang2023dawn} --- have been successful in specific task domains. However, whether these initial successes will translate to robust performance across the diverse  landscape of real-world visual reasoning tasks remains an open question.
We extend this work in two ways: First, we evaluate a tool-augmented VLM across all three task domains to assess whether external tool use can overcome serial processing deficits. Second, we propose training regimes that could instill more generalizable serial visual reasoning capabilities in VLMs. 

\section{Test Domain 1: Geometric Concepts}
The first domain in which we investigate a potential serial processing deficit in VLMs is geometric reasoning. A long tradition in cognitive science has examined how humans reason over geometric patterns, demonstrating that response times (RTs) increase systematically with the structural complexity of the stimulus \citep{just1985cognitive}. In particular, a growing body of work has argued that humans interpret geometric figures by inferring the underlying generative rules that produced them, rather than merely matching surface features. This \textit{program induction} approach has been explored in experiments that manipulate the minimum description length (MDL) of symbolic programs used to generate visual stimuli -- where MDL reflects the number of discrete compositional steps required to construct the pattern from a fixed vocabulary of geometric primitives and relations \citep{SABLEMEYER2022101527}. 
Such studies suggest that human geometric reasoning is fundamentally compositional and serial: more complex patterns require more sequential operations, resulting in longer RTs but typically preserved accuracy. Recent work introduced the \textit{Geoclidean} domain-specific language (DSL) to generate 2D geometric stimuli by composing geometric primitives \cite{hsu2022geoclidean}. This work revealed that while humans are adept at inferring abstract geometric rules, standard transformer and CNN-based vision encoders are insensitive to this structure. However, it remains unclear whether modern VLMs, which incorporate language-based reasoning, exhibit similar deficits --- particularly for complex concepts that demand greater serial processing in humans. We test this by parametrically varying the complexity of geometric concepts generated using the Geoclidean DSL, predicting that model performance will degrade with increasing complexity, even as human accuracy remains stable.

\begin{figure}[h]
\begin{center}
\includegraphics[width=1\linewidth, trim = 0 0 0 50, clip]{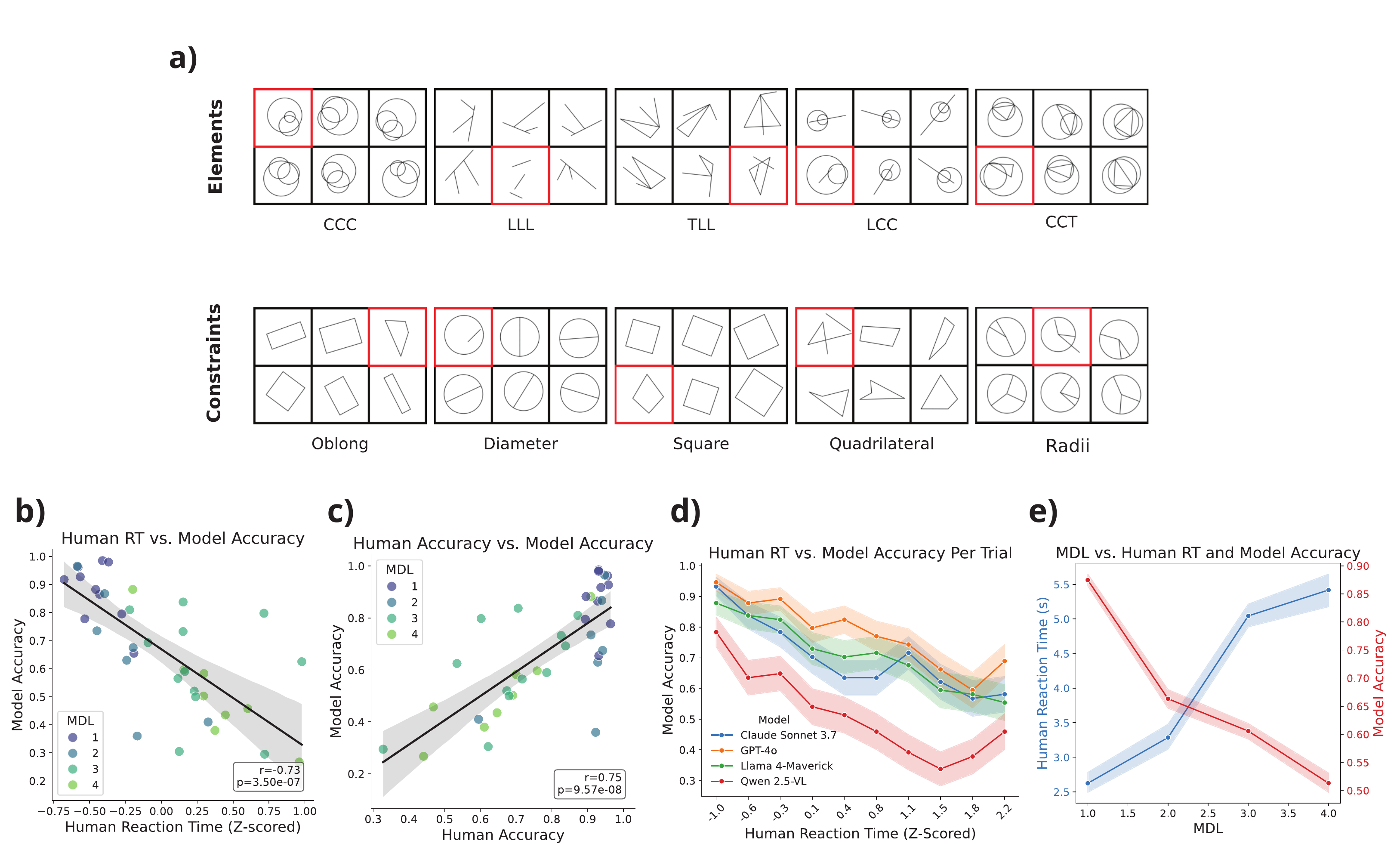}
\end{center}
  \caption{\textbf{Geometric Reasoning Task.}
\textbf{a)} Example oddball detection trials from a subset of 37 geometric concepts spanning both primitive elements (left) and relational constraints (right), generated with Geoclidean DSL. 
\textbf{b)} Relationship between z-scored human reaction time and model accuracy; each point is one geometric concept and color denotes program complexity (Minimum Description Length, MDL). 
\textbf{c)} Correlation between human and model accuracy across concepts. 
\textbf{d)} Trial-level correlation between human reaction time (RT) and model accuracy. 
\textbf{e)} Human RT (blue, left axis) and model accuracy (red, right axis) as a function of MDL. Shaded regions are 95\% confidence intervals.}
  \label{fig:geoclidean}
\end{figure}

\subsection{Methods}
To evaluate the sensitivity of VLMs to geometric concepts, we adapted the ``oddball'' detection paradigm from foundational studies of non-verbal human geometric intuition \citep{dehaene2006core}. Each trial consisted of an array of six images. Five of these images consistently instantiated a particular geometric concept, while the sixth item—the 'oddball'—violated this concept. We evaluated model and human performance on a geometric oddball task generated using the Geoclidean DSL \cite{hsu2022geoclidean}. Stimuli from 37 distinct concepts belonging to two categories --- Geoclidean-Elements and Geoclidean-Constraints --- were generated. (Figure~\ref{fig:geoclidean}a). These concepts spanned a range of geometric properties, including Euclidean geometry (e.g., parallelism, right angles, line bisecting), specific figures (e.g., squares, triangles, polygons), and constraints and relations (tangency, midpoint relations, angle constraints). 
We created 100 trials for each of the 37 concepts, yielding a total of 3,700 trials. Human participants each completed a randomly selected subset of 50 trials, drawn from a uniformly sampled 20\% subset of the model evaluation trials. These trials were sampled such that each task condition was equally represented, and each trial received at least 20 independent human judgments. The oddball was constructed by removing two relational constraints from the reference concept while preserving its underlying components. Concepts ranged in program complexity from MDL 1 to 4, where Minimum Description Length (MDL) denotes the number of geometric primitives needed to generate the concept.
Human participants and VLMs were tasked with identifying the position of the oddball within the array. (Appendix Figure~\ref{fig:geo_instruction}). Human participants indicated their choice by clicking on the target image, and their reaction times were recorded. (Appendix Figure~\ref{fig:geo_example}). VLMs were presented with the same array of images, each overlaid with a red index in the corner, and prompted to output the index corresponding to the oddball. (Appendix~\ref{appendix:geoclidean_prompt}). We evaluated four large multimodal models --- GPT-4o \citep{achiam2023gpt}, Claude Sonnet 3.7 \citep{anthropic2023claude}, Llama 4 Maverick \citep{meta2024llama4}, Qwen2.5 VL \citep{bai2025qwen25vltechnicalreport} --- on this task.

\subsection{Results}
We first evaluated the alignment between human and model performance across geometric concepts. As shown in Figure~\ref{fig:geoclidean}c, human accuracy and model accuracy were positively correlated across concepts ($r = 0.75$, $p = 9.57 \times 10^{-8}$), suggesting that both humans and vision-language models (VLMs) are broadly sensitive to task difficulty -- though potentially for different underlying reasons.

Importantly, we also observed an inverse relationship between human reaction time (RT) and model accuracy (Figure~\ref{fig:geoclidean}b). Concepts that took humans longer to solve were precisely those where models performed worst ($r = -0.73$, $p = 3.50 \times 10^{-7}$), implying that model failures are most acute on tasks requiring extended human deliberation. Moreover, at the level of trials, we observed a similar inverse relationship between human RT and model accuracy (Figure~\ref{fig:geoclidean}d).
We formalized this pattern by analyzing performance as a function of minimum description length (MDL). As shown in Figure~\ref{fig:geoclidean}e, human RT increased with MDL (blue), while model accuracy declined (red). This divergence suggests that humans engage more cognitive effort as complexity rises, whereas models lack the structured inference mechanisms necessary to handle deeper compositional programs.

Together, these findings support our serial processing deficit hypothesis: while humans accommodate complexity via slower, sequential reasoning, VLMs exhibit declining accuracy as task complexity increases -- despite equivalent visual input. In our subsequent experiments we show that similar patterns can be observed in other settings where serial processing is important.

\begin{figure}[t]
  \begin{center}
  \includegraphics[width=0.8\linewidth,trim = 0 10 0 10, clip]{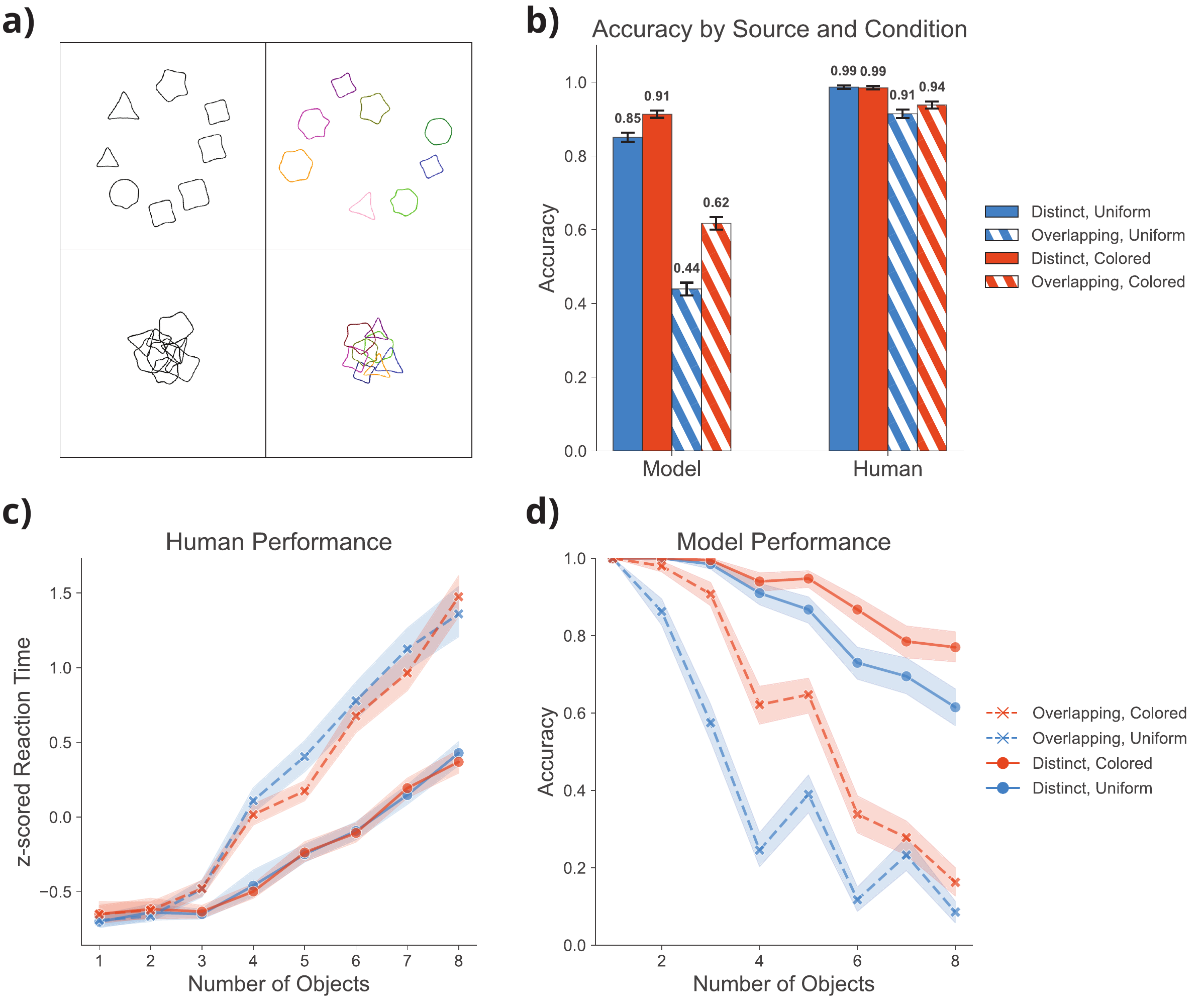}
  \end{center}
  \caption{\textbf{Numerosity Task.}
\textbf{a)} Example stimuli from the four experimental conditions: non-overlapping uniformly colored (top-left), non-overlapping uniquely colored (top-right), overlapping uniformly colored (bottom-left), and overlapping uniquely colored (bottom-right). 
\textbf{b)} Model accuracy as a function of numerosity across the four conditions. 
\textbf{c)} Mean accuracy for humans and models across each condition, averaged across numerosities. 
\textbf{d)} Human z-scored reaction times as a function of numerosity across the four conditions. Shaded regions indicate 95\% confidence intervals.}
  \label{fig:numerosity}
\end{figure}

%   \vspace{0.5em}
%   \noindent
%   \textbf{Legend:} 
% \textcolor{orange}{\rule{1.2em}{0.6ex}} colored \quad
% \textcolor{blue}{\rule{1.2em}{0.6ex}} uniform \quad
% \tikz[baseline=-0.5ex] \draw[thick] (0,0) -- (1.5em,0); non-overlapping \quad
% \tikz[baseline=-0.5ex] \draw[dashed, thick] (0,0) -- (1.5em,0); overlapping \quad \newline
% \textcolor{black}{\texttt{///}} overlapping (hashed bar, panel c)

\section{Test Domain 2: Numerical Estimation}

Numerical estimation --- such as assessing how many objects appear in a scene --- is a classic serial processing task. Human numerical estimation abilities are characterized by a transition between the near-instantaneous recognition of small quantities (up to 4–6 items), a process known as subitization, and the serial enumeration of larger quantities \citep{trick1994small, kaufman1949discrimination}. As humans enumerate larger sets or when visual discrimination becomes challenging, we  deploy serial attention to identify each object --- binding features to locations and maintaining distinct object representations \citep{mazza2015multiple, xu2009selecting}. Such serial attention-based enumeration becomes more challenging as objects become harder to distinguish \citep{franconeri2013flexible, lavie2005distracted}. Prior work has documented VLMs' difficulty in counting  \citep{pothiraj2025captureevaluatingspatialreasoning,campbell2025understandinglimitsvisionlanguage}; we predict that, in alignment with human behavior, VLM performance will be highest when the number of objects is low and the objects are non-overlapping. However, while humans maintain accuracy in challenging conditions \citep{vetter2008modulating} (e.g. overlapping objects at higher numerosities) by deploying serial attention, we hypothesize that VLM performance will degrade rapidly. Furthermore, we predict that this degradation in performance would coincide with an increase in human reaction time. Critically, we anticipate an asymmetry in performance, where VLM performance increases when overlapping objects are uniquely colored compared to when they are uniformly colored, a pattern not mirrored in human performance. This prediction stems from our hypothesis that VLMs' serialization capabilities depend heavily on their ability to ground reasoning in their linguistic chain of thought, which is enabled when the objects can be uniquely identified.% by their color in context. 

\subsection{Methods}
We evaluated model and human performance on a visual enumeration task designed to systematically vary perceptual individuation load. Stimuli consisted of 2-dimensional shapes generated using Hermitian splines, featuring between 3 and 5 pointed contours with randomly jittered fixed points (Figure~\ref{fig:numerosity}a).
To parametrically manipulate serial processing demands, we varied both numerosity (1-8 objects) and object distinctiveness across two dimensions: spatial arrangement (overlapping vs. non-overlapping) and color (uniformly colored vs. uniquely colored). This factorial design created four distinct conditions: (1) non-overlapping, uniformly colored objects; (2) non-overlapping, uniquely colored objects; (3) overlapping, uniformly colored objects; and (4) overlapping, uniquely colored objects. These manipulations were designed to vary reliance on serial attention required for object identification while controlling for the total number of objects.

In the uniform color conditions, all objects had the same hue, requiring individuation based solely on their spatial arrangement. In the unique color conditions, each object was assigned a distinct hue, providing an additional dimension for differentiation. In the non-overlapping conditions, objects were positioned with clear spatial separation, while in the overlapping conditions, objects were arranged with 60-80\% intersection between adjacent shapes so boundary resolution required attention.

Each of the four conditions was instantiated across 100 trials for each numerosity level (1-8), yielding 3,200 total examples. Each human participant completed a randomly selected set of 50 trials, drawn from a uniformly sampled 20\% of the model evaluation set. Sampling was stratified to ensure balanced representation across task conditions, with each trial evaluated by at least 10 independent participants.
Human participants and VLMs were tasked with identifying the number of objects in the scene (Appendix Figure~\ref{fig:num_instruction}).
Participants performed the task while reaction times and accuracy were recorded (Appendix Figure~\ref{fig:num_example}). VLMs were prompted to describe the image and report the number of objects present (Appendix~\ref{appendix:numerosity_prompt}). This allowed us to directly measure the impact of increasing perceptual individuation load on task performance, and to compare the results for humans and VLMs. 

\subsection{Results}

We first evaluated the alignment between human and model performance across the four task conditions. As shown in Figure~\ref{fig:numerosity}c, humans sustained accuracy across conditions, while model accuracy significantly degraded in the overlapping conditions. 

We also observed an inverse relationship between model performance and numerosity (Figure~\ref{fig:numerosity}b), indicating that models struggled both with high numerosities and perceptual individual loads. These same conditions took humans longer to solve ($r = -0.97$, $p = 8.22 \times 10^{-5}$) and we observed an inverse relationship between human RT and model performance.%, implying that model failures are most acute on tasks requiring extended human deliberation. 
Performance was also consistently impaired on overlapping trials, with models showing substantially lower accuracy ($t=-8.43$, $p=7.9\times 10^{-10}$), and humans requiring more time ($t=3.45$, $p=0.005$) and exhibiting reduced accuracy ($t=-2.43$, $p=0.023$).  

Finally, we observed that model performance was significantly better in the colored overlapping condition compared with the uniform overlapping condition ($t=5.90$, $p=8.1\times 10^{-7}$; Figure~\ref{fig:numerosity}b). This model-specific improvement was not accompanied by a corresponding decrease in human reaction time ($t=-1.17$, $p=0.28$; Figure~\ref{fig:numerosity}d) or accuracy ($t=2.06$, $p=0.078$), suggesting that color aided model performance not by reducing perceptual difficulty, but by facilitating object individuation. This supports our hypothesis that VLM serialization is improved when objects can be uniquely referenced by their color in context, a  capability that humans do not benefit from.

\begin{figure}[t]
  \begin{center}
  \includegraphics[width=\linewidth]{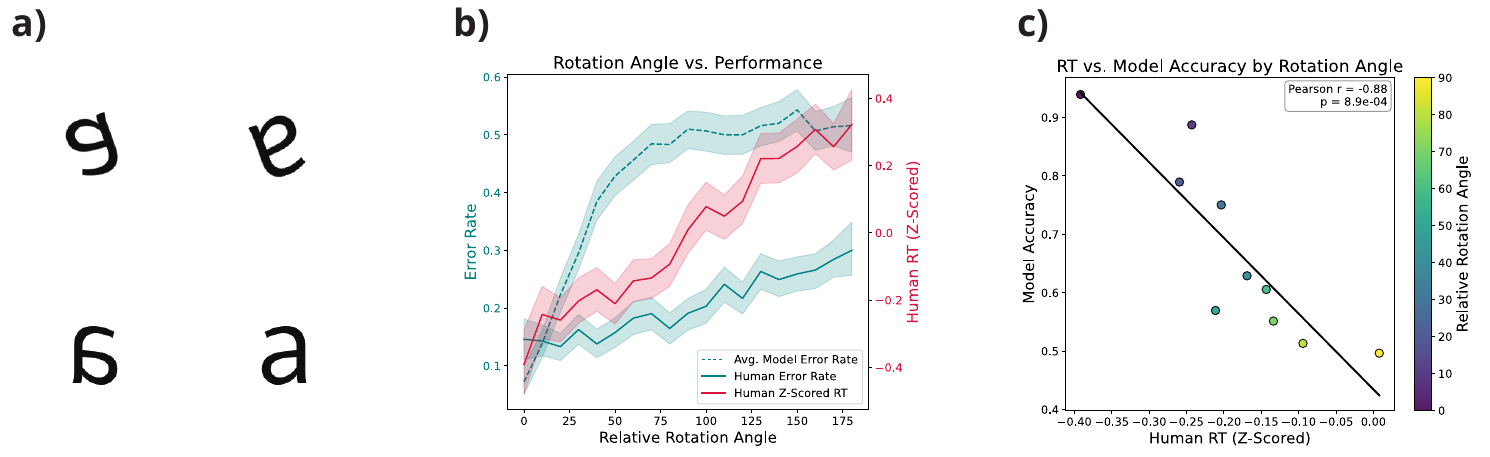}
  \end{center}
  \caption{\textbf{Mental Rotation Task.}
    \textbf{a)} Example stimuli from the mental rotation task. Participants and models judged whether a rotated letter was the same or a mirror-reversed version of a reference. 
    \textbf{b)} Human error rate (solid line), model error rate (dashed line), and z-scored human reaction time (right axis) plotted against relative rotation angle. Shaded regions denote 95\% confidence intervals.
    \textbf{c)} Relationship between z-scored human reaction time and model accuracy across rotation angles ($\leq 90^{\circ}$). Each point corresponds to a single rotation angle; colors indicate relative angular disparity.}
    \label{fig:rotation}
\end{figure}

\section{Test Domain 3: Mental Rotation}

Mental rotation provides a final test case where the sequential nature of the task cannot be effectively substituted with linguistic reasoning. Unlike symbolic pattern recognition, mental rotation requires the manipulation of spatial representations through continuous transformations---operations that are difficult to express in discrete, language-based tokens.
A standard mental rotation task presents participants with two images which are either two distinct objects or one object shown from different angles. Decades of cognitive science research demonstrate that humans solve these tasks by mentally rotating one object to align with another, an analog process captured by the robust linear relationship between angular disparity and reaction time \cite{shepard1971mental, cooper1973chronometric}. Functional imaging confirms that this process involves incremental spatial adjustments, engaging parietal regions known to support visuospatial transformation \cite{zacks1999imagined, just1985cognitive}.
These transformations rely on what \cite{kosslyn1994image} termed “depictive representations” -- mental images that preserve geometric structure and spatial relations. Because such analog operations are difficult to serialize into linguistic prompts, mental rotation offers a particularly stringent test of whether VLMs can support visual reasoning that is not scaffolded by language.

\subsection{Methods}

We evaluated human and model performance on a classic mental rotation task, adapted from prior cognitive science paradigms \citep{shepard1971mental, cooper1973chronometric}. Stimuli consisted of character-like shapes that were either identical or mirror-reversed versions of a reference, displayed at various rotation angles (Figure~\ref{fig:rotation}a). Each trial presented a pair of shapes, and the task was to determine whether they matched or were mirrored. (Appendix Figure~\ref{fig:rotation_instruction}).

Rotation angles varied systematically from 0\textdegree{} to 360\textdegree{} in 10\textdegree{} increments, yielding 36 distinct orientations. For each angle and letter stimulus (26 total), we generated four trial types crossing two factors: whether the pair was the same or mirror-reversed, and whether the first image was mirrored or not. This design resulted in $36 \times 4 \times 26 = 3744$ total trials. Trial order was fully randomized, and conditions were balanced across same/mirror status and initial image mirroring. As before, each participant completed a randomly selected set of 50 trials, drawn from a uniformly sampled 20\% subset of the full model evaluation set. Sampling ensured equal representation across task conditions, and each trial was judged by at least 10 independent human participants.

Human participants completed the task via mouse click, and RT was recorded from stimulus onset to response. (Appendix Figure~\ref{fig:rotation_example}). VLMs were shown images of the two shapes, and prompted to indicate whether the shapes are ‘same’ or ‘different’. (Appendix~\ref{appendix:rotation_prompt}). Model performance was evaluated as overall accuracy across each numerosity and load condition. 

\subsection{Results}

We first examined the effect of rotation angle on  performance. As shown in Figure~\ref{fig:rotation}b, human reaction time increased monotonically with angular disparity, replicating the classic linear RT-angle relationship. This trend reflects the analog nature of human mental rotation: more extreme angles require longer sequences of transformations \citep{just1985cognitive, zacks1999imagined}.

As predicted, VLM performance exhibited an inverse relationship. As rotation angle increased, model accuracy declined, with particularly sharp drops in accuracy beyond 75\textdegree, consistent with prior findings on spatial reasoning limitations in vision-language systems \citep{stogiannidis2025mind, chen2025why}. This decline is reflected in the average model error rate (dashed line, Figure~\ref{fig:rotation}b), which rose substantially across angle bins.

Human accuracy also declined with rotation angle, but more modestly. While participants made more errors at higher angles, their performance degraded far less steeply than that of the models, indicating partial robustness to increased serial demands. This distinction reinforces the idea that humans can compensate for increasing task difficulty by allocating more cognitive resources---reflected in longer RTs---whereas VLMs currently lack such dynamic inference capabilities.

Critically, we observed a strong negative correlation between human RT and model accuracy across rotation angles (Figure~\ref{fig:rotation}c; $r= -0.88$, $p = 8.9 \times 10^{-4}$). Trials that demanded longer processing times from humans were precisely those where VLMs performed worst. This inverse relationship supports the serial processing deficit hypothesis: the tasks that elicit deeper, stepwise transformations in humans are the ones that reveal structural limitations in current VLMs.

Together, these findings suggest that mental rotation engages a form of non-linguistic serial reasoning that vision-language models do not replicate. Despite identical visual input, models falter where humans invoke continuous internal transformations -- highlighting a fundamental divergence in the underlying mechanisms of reasoning.

\begin{figure}[ht]
  \begin{center}
  \includegraphics[width=\linewidth]{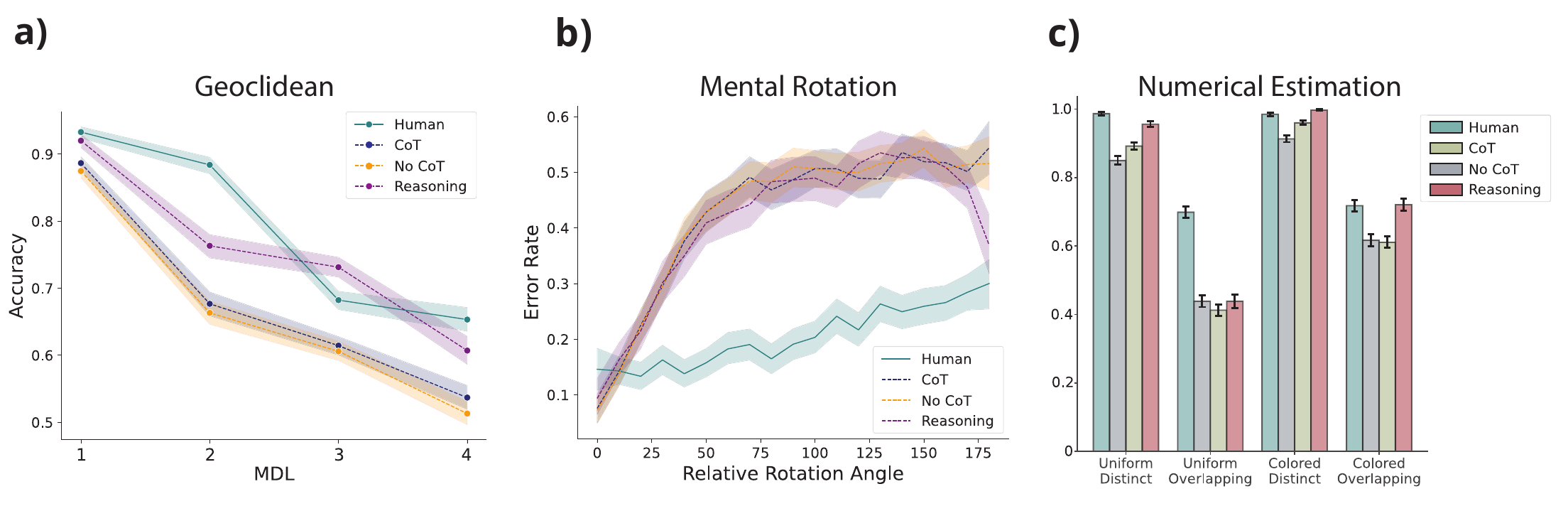}
  \end{center}
  \caption{\textbf{Results for Augmented VLMs.}
    \textbf{a)} Accuracy as a function of MDL for humans and different model conditions. 
    \textbf{b)} Error rate as a function of relative rotation angle for humans (green line) and models under different augmentation paradigms. 
    \textbf{c)} Mean accuracy for humans and models across each condition, averaged across numerosities. 
    \textbf{d)} Accuracy on a challenging subset of Geoclidean and rotation tasks, comparing humans, reasoning-augmented models, and GPT-o3 with tool use. 
    \textbf{e)} Accuracy by condition for $n = 8$ items, across humans, reasoning models, and GPT-o3 with tool use. 
    Shaded regions in \textbf{a)} and \textbf{b)} represent 95\% confidence intervals.
    }
    \label{fig:causal_analysis}
\end{figure}

\section{Causal evidence for a serial processing deficit}

One way to causally test the hypothesis that VLMs lack serial processing capabilities is to augment VLMs with a form of serial processing, and test whether this enables them to solve more complex visual tasks. To investigate this, we tested models augmented with three forms of serial processing: Chain-of-thought (CoT), reasoning training, and tool use (e.g., the ability to crop and rotate images).% to focus on particular regions). 

We do not expect these forms of serial processing to be effective in all tasks. Improvements from CoT and reasoning training should be limited to cases where individual objects can be uniquely referenced in the models CoT. Our numerosity estimation task provides an ideal setting to test this prediction. In this task, when objects are both non-overlapping and distinctly colored, textual CoT can effectively be used to describe the objects one at a time (based on color), whereas textual CoT should be less effective when objects are overlapping or uniform in color. Consistent with this, we see that this condition (colored distinct) is the one where CoT ($t = 7.79$,$p < 0.005$) and reasoning training ($t = 13.81$, $p < 0.005$) help the most. Reasoning training results in nearly perfect performance in this condition (99.7\% accuracy; Figure \ref{fig:causal_analysis}), but does not improve performance significantly in the overlapping conditions ($p > 0.5$).

Similarly, improvements from tool use should be limited to cases where objects can be isolated by cropping an image. This results in near-perfect accuracy on our rotation task (Table \ref{table:tool-use-results}, $p < 0.05$) and perfect performance on conditions in which the objects are spatially separated (uniform distinct and colored distinct) in our numerosity task (100\% accuracy, $p < 0.005$). However, tool use does not improve on baseline performance for either of the overlapping conditions (Table \ref{table:tool-use-results}, $p > 0.1$). 

Two major conclusions can be drawn from these results. First, they show that augmenting VLMs with serial processing enables them to solve visually complex tasks, thus providing causal evidence for the serial processing deficit hypothesis. Second, these results highlight the limitations of existing methods for serial processing in VLMs (textual CoT and tool use) as their application is limited to specific settings. This points to the need for developing intrinsically visual forms of serial processing to help VLMs deal with a broader range of visually complex tasks.

\section{Discussion}
\label{section:discussion}

We have shown that Vision-Language Models (VLMs) struggle with tasks that require visually grounded serial processing, revealing behavioral limitations that parallel specific constraints in human cognition. These results offer an explanatory account of why models so adept at many tasks struggle with activities as simple as counting. They also point toward concrete strategies for improving VLM performance by targeting improvements in visual serial processing.
Across three experimental paradigms, we observed a consistent pattern: VLM performance inversely correlates with human reaction time. This relationship provides strong evidence for our serial processing deficit hypothesis, suggesting that operations requiring sequential processing in humans---indicated by longer reaction times---correspond precisely to operations that current VLMs fail to perform. The cross-domain consistency of this relationship indicates that serial processing capacity represents a general computational constraint rather than a domain-specific limitation.

In geometric reasoning, as program complexity (measured by MDL) increased, human reaction times rose, whereas VLM performance deteriorated. Previous work using similar tasks argued that human performance implies symbolic program induction, a capability that neural network models lack \cite{SABLEMEYER2022101527}. Contrary to this finding, we show that neural networks (here, VLMs) \textit{do} correlate with both human behavior and measures of task complexity. It is an open question whether to interpret this as evidence that VLMs perform program induction, or that the human behavioral measures do not reflect  program induction. However, our results indicate that VLMs appear unable to flexibly compose visual operations in ways that scale with program complexity. 

The numerical estimation task revealed that VLMs demonstrate subitizing-like behavior for small quantities under optimal viewing conditions but show performance degradation when object individuation becomes challenging. Humans maintain accuracy in these conditions by deploying serial attention---binding features to locations and maintaining distinct object representations---at the cost of increased processing time \cite{mazza2015multiple, xu2009selecting}. We also find that VLMs benefit from unique coloring in overlapping object conditions, while humans show no comparable advantage. This contrasts with some findings, such as those by \cite{rahmanzadehgervi2025visionlanguagemodelsblind}, who reported that coloring conditions in numerosity tasks made no significant difference for most of the VLMs tested. Here, we show that when objects overlap, unique coloring provides a distinct and significant advantage to VLMs. This advantage likely arises because distinct colors allow VLMs to individually reference and effectively count objects, thereby enabling them to linguistically ground features and perform feature binding in a way that is not possible in uniformly colored conditions. 

Mental rotation provided a particularly stringent test of \textit{visual} serial processing, as this analog transformation is difficult to decompose into discrete linguistic steps. While humans show a linear relationship between rotation angle and reaction time \cite{shepard1971mental} (reflecting an incremental transformation process), and relatively limited effects of rotation angle on accuracy, VLMs exhibited sharp performance drops as rotation angles increased. This suggests a limitation in performing the continuous transformations that humans execute through depictive representations \cite{kosslyn1994image}.
Importantly, we are not claiming that models are incapable of serial processing---indeed, autoregressive generation implicitly implements serial processing. However, this seems to be a \textit{strictly linguistic} form of serialization that is difficult to apply to visual tasks like mental rotation.

\subsection{Architectural Improvements}

Recent advances in VLMs, such as V* \citep{yang2023dawn} and tool-augmented models, are intended to mimic how humans process scenes during challenging visual reasoning tasks. Rather than processing entire images at once, humans sequentially focus their attention on different regions, using high-resolution central vision to examine each area in detail—a strategy that helps overcome the limitations of trying to process all visual information simultaneously \citep{stewart2020review}.
Our work shows two fundamental limitations of these current augmentations. First, linguistic Chain-of-Thought (via prompting and reasoning fine-tuning) fails when input cannot be grounded in language. Second, while image tool use effectively solves well-defined transformations with explicit algorithms, it struggles with tasks requiring more general forms of serial reasoning that do not involve explicit transformations of the input images.

A more promising path forward involves inducing a more sequential and deliberate mode of processing directly, with a new class of models trained using Visually Grounded Reinforcement Learning (RL) showing particular promise. Rather than processing an image in a single pass, methods like ViGoRL, GRIT, UniVG-R1, and Ground-R1 use RL to train a policy that generates an explicit reasoning trace where each textual thought is anchored to a specific coordinate or bounding box in the visual input \citep{sarch2025vigorl, wang2025grit, li2025univgr1, cao2025groundr1incentivizinggroundedvisual}. This process of generating a sequence of region-grounded thoughts functions as a computational analogue to the human saccade-and-fixate cycle, forcing the model to abandon its default holistic strategy and adopt a more structured, pseudo-serial approach to scene analysis. This also increases model interpretability, given the visual-attentional path to an answer can be explicitly examined.

Other post-training methods inspired by human serial vision show promise in specific domains. Molmo \citep{deitke2024molmo} demonstrates improved performance on counting when trained to successively point to objects. Similar fine-tuning approaches for sequential attention could prove beneficial in other visual reasoning domains. The processing deficits we observed emerge from interference between visual elements, as shown in our numerosity experiments and by \citep{campbell2025understandinglimitsvisionlanguage}. This interference could be mitigated through causal masking around fixation points identified by models like Molmo, producing effects similar to the human distinction between foveal and peripheral vision \citep{stewart2020review}. This controlled, sequential attention could reduce the interference we observed and provide a  path toward more human-like serial visual processing.

\subsection{Limitations \& Future Directions}

This work has several limitations. First, our analysis is constrained to a limited set of VLMs. We chose models that are competitive with other available models, but our selection does not necessarily capture the full diversity of performance. Second, our tasks were selected to manipulate serial processing load in visual reasoning; future work could consider a broader range of tasks. Specifically, extending our analysis to tasks like visual search and spatial reasoning \cite{gilden2010serial, reimer2019serialsearch} would increase the generality of our results. Moreover, evaluating spatial reasoning alongside a broader range of tasks with varying serial processing load may allow us to more effectively dissociate spatial reasoning deficits from serial reasoning deficits.

\subsection{Conclusion}

Our work provides a unifying explanation for the discrepancy between VLMs' impressive benchmark performance and their difficulty with seemingly simple visual reasoning tasks. The serial processing deficit hypothesis offers both a theoretical framework for understanding current model limitations and a roadmap for future architectural innovations.
By identifying specific visual reasoning operations where VLMs diverge from human behavior, we contribute to the understanding of computational principles underlying visual intelligence. As the field develops the next generation of visual reasoning models, addressing the serial processing deficit may be key to achieving more human-like flexibility across diverse visual tasks---combining the parallel pattern recognition capabilities of contemporary VLMs with the serial reasoning characteristics that have been a hallmark of human visual cognition.

\newpage

\section{Ethics Statement}

We conducted our human behavioral experiments under an IRB-approved protocol. Further information, such as about the potential risks of behavioral data collection and details about compensation, can be found in the Appendix. 

We address potential societal impacts of our work in the Discussion. Our experiments do not pose a high risk for misuse since we do not release new models or use any proprietary assets. 

\section{Reproducibility Statement}

All experimental details and methods for testing models are described in the main text and Appendix. The code base necessary to generate experimental stimuli, run model experiments, and human behavioral experiments will be open-sourced upon publication. We will include our full datasets and instructions to faithfully reproduce our work. 

We include screenshots of instructions provided to participants in the Appendix, as well as details on compute needed to run model experiments. Our work does not involve any novel theory that would require formal justification. Finally, we report statistical significance for all results where appropriate in the main text and Appendix. 

% \section*{References}

\bibliography{ref}
\bibliographystyle{plainnat}

\appendix
\section{Appendix: Prompts for Vision-Language Model Experiments}
\label{app:prompts}

\lstnewenvironment{prompt}
  {\lstset{
    basicstyle=\ttfamily\scriptsize, % Use small typewriter font
    breaklines=true,            % Automatic line breaking
    breakatwhitespace=false,    % Break lines not only at whitespaces
    frame=leftline,             % Line at the left side (optional)
    breakindent=10pt,
    framesep=5pt,               % Distance of text to the frame (optional)
    xleftmargin=5pt,            % Indent the listing environment (optional)
    aboveskip=10pt,             % Space above the listing (optional)
    belowskip=10pt,             % Space below the listing (optional)
  }}
  {}

\subsection{Geoclidean}

\subsubsection{Baseline Prompt}

\begin{prompt}
Which of the 6 shapes is different from the others? 

Return only the number of the odd one out in square brackets (e.g., [3]) without any additional reasoning or justification.
\end{prompt}
\label{appendix:geoclidean_prompt}

\subsubsection{Chain-of-Thought Prompt}
\begin{prompt}
Which of the 6 shapes is different from the others?

Analyze the relations between the parts of each stimulus to identify the stimulus that is different from the rest. Return your reasoning followed by the integer label of the odd one out enclosed in square brackets (e.g., [3]).
\end{prompt}
\label{appendix:geoclidean_CoT_prompt}

\subsection{Rotation}
\subsubsection{Baseline Prompt}

\begin{prompt}
You will be shown two objects that are rotated at different angles. Your task is to determine whether these objects are:

A) The **SAME** object shown at different rotations
B) **DIFFERENT** objects that are mirror images of each other (these mirror images may also be rotated at different angles)

Even when rotated, identical objects will have the same features in the same arrangement. Mirror images will have reversed features that cannot be matched by rotation alone.

## Response Format:
[1] - The objects are identical but rotated differently
[0] - The objects are mirror images of each other

Please just include your final answer in the format [1] or [0] without any additional reasoning or justification.
\end{prompt}
\label{appendix:rotation_prompt}

\subsubsection{Chain-of-Thought Prompt}

\begin{prompt}
You will be shown two objects that are rotated at different angles. Your task is to determine whether these objects are:

A) The **SAME** object shown at different rotations
B) **DIFFERENT** objects that are mirror images of each other (these mirror images may also be rotated at different angles)

Even when rotated, identical objects will have the same features in the same arrangement. Mirror images will have reversed features that cannot be matched by rotation alone.

## Response Format:
[1] - The objects are identical but rotated differently
[0] - The objects are mirror images of each other

Please include your reasoning followed by your final answer in the format [1] or [0].
\end{prompt}
\label{appendix:rotation_CoT_prompt}

\subsection{Numerosity}

\subsubsection{Baseline Prompt}
\begin{prompt}
The following images contains 1 or more objects that may or may not be overlapping.

Count the total number of objects in the scene, and return your final answer as an integer enclosed in square brackets (e.g., [10]).
\end{prompt}
\label{appendix:numerosity_prompt}

\subsubsection{Chain-of-Thought Prompt}

\begin{prompt}
The following images contain 1 or more objects that may or may not be overlapping.

Count the total number of objects in the scene one at a time. Once you have described all objects, return your final answer as an integer enclosed in square brackets (e.g., [10]).
\end{prompt}
\label{appendix:numerosity_CoT_prompt}

\section{Appendix: Individual Model Results}

\begin{figure}[h]
  \begin{center}
  \includegraphics[width=\linewidth]{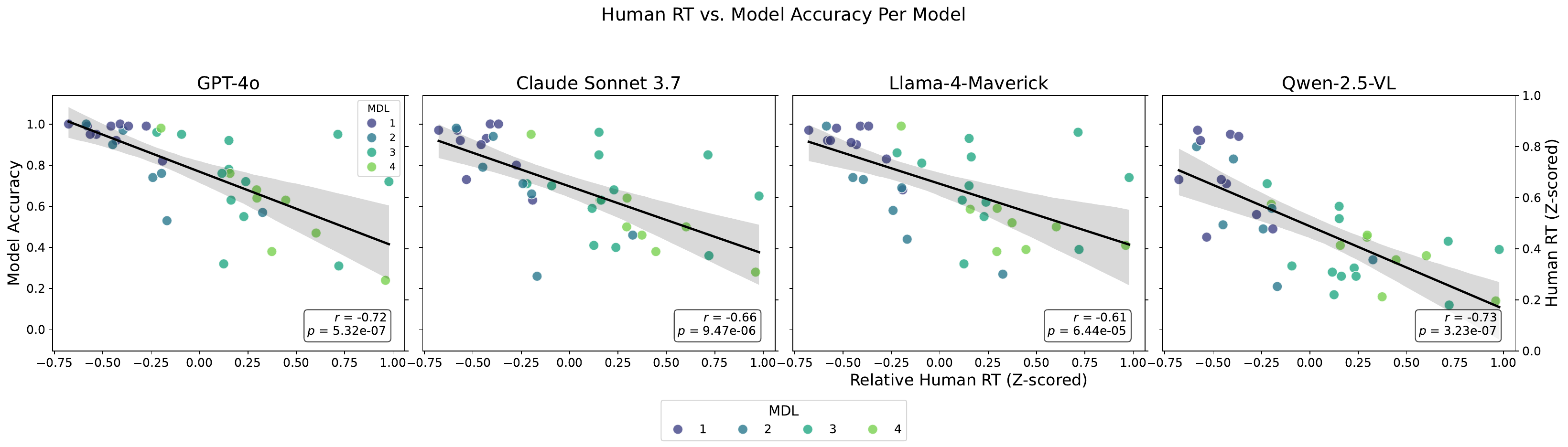}
  \end{center}
  \caption{\textbf{Human RT vs. Model Accuracy by Model.}
Z-scored human reaction time and model accuracy plotted across geometric concepts, shown separately for each VLM. Each point corresponds to one concept, colored by program complexity (MDL). Black lines show linear regression fits with 95\% confidence intervals.}
    \label{fig:geoclidean_individualmodels_figb}
\end{figure}

\begin{figure}[h]
  \begin{center}
  \includegraphics[width=\linewidth]{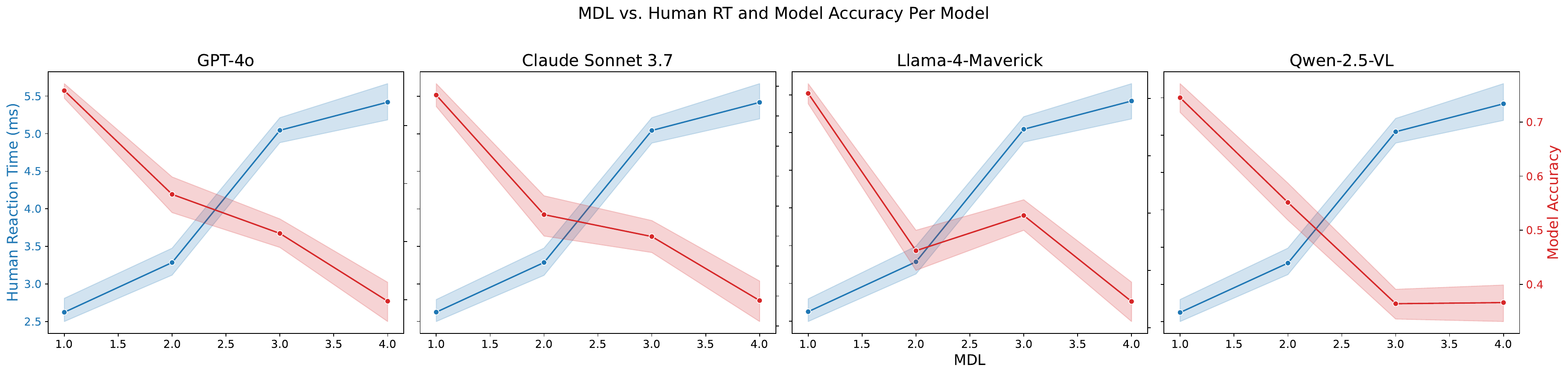}
  \end{center}
  \caption{\textbf{MDL vs. Human RT and Model Accuracy by Model.} Human reaction time (blue, left axis) and model accuracy (red, right axis) plotted as a function of MDL, shown separately for each VLM. Shaded regions indicate 95\% confidence intervals.}
    \label{fig:geoclidean_individualmodels_fige}
\end{figure}

\newpage

\begin{figure}[h]
  \begin{center}
  \includegraphics[width=\linewidth]{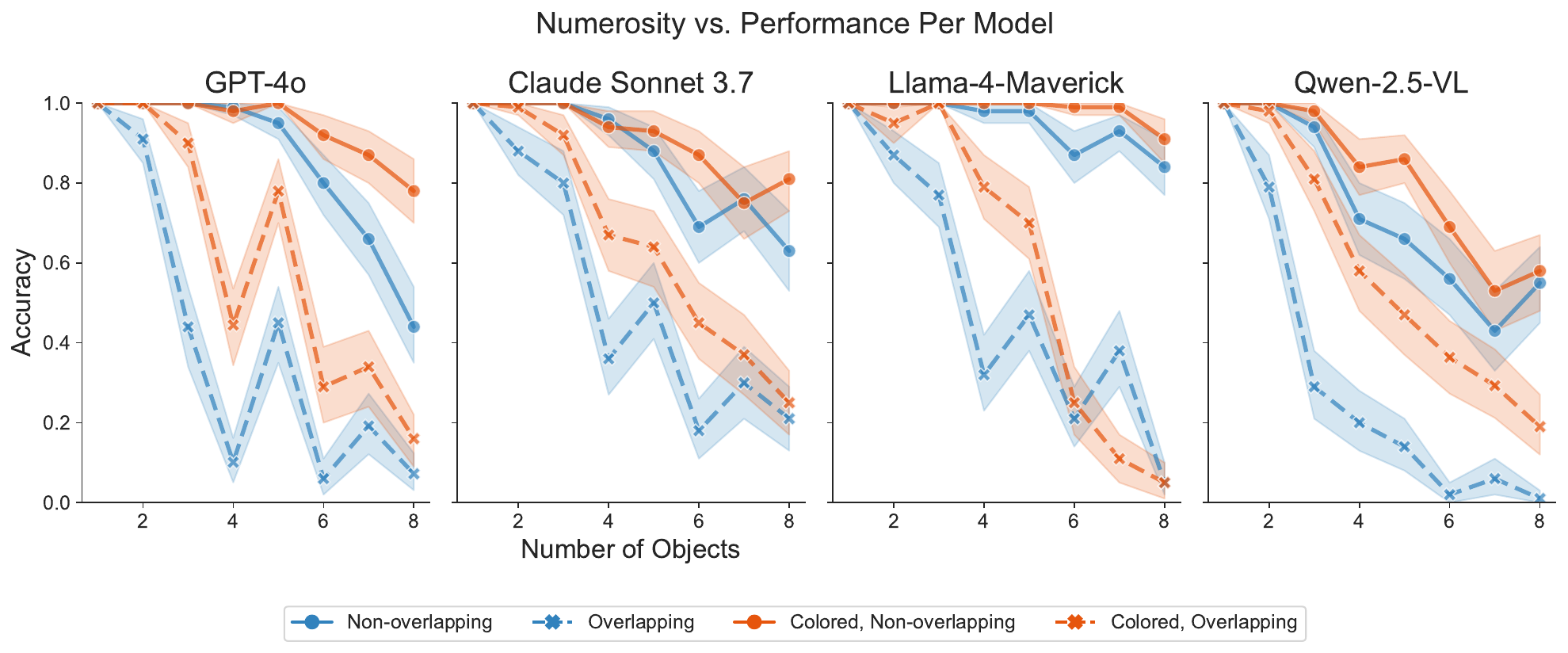}
  \end{center}
  \caption{\textbf{Numerosity vs. Performance by Model.}
Model accuracy in the overlapping condition (dashed), distinct condition (solid), colored condition (orange), and uniform condition (blue) plotted as a function of the number of objects in the scene, shown separately for each VLM. Shaded regions indicate 95\% confidence intervals.}
    \label{fig:numerosity_individualmodels}
\end{figure}

\begin{figure}[h]
  \begin{center}
  \includegraphics[width=\linewidth]{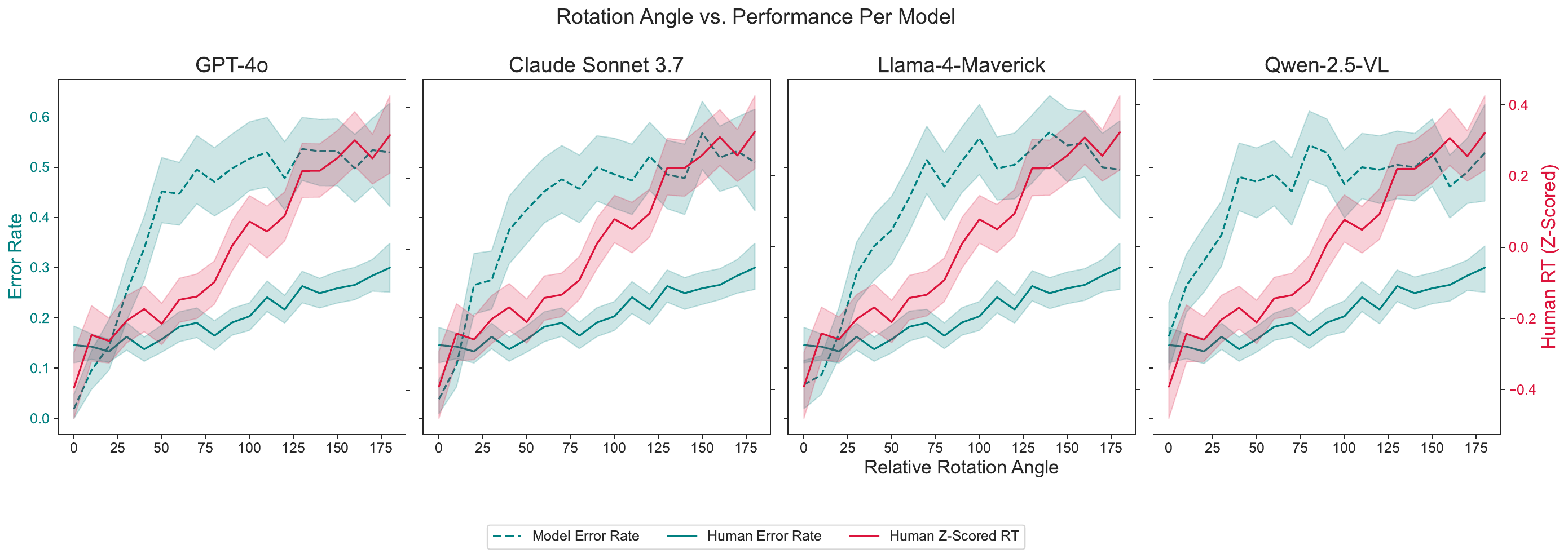}
  \end{center}
  \caption{\textbf{Rotation Angle vs. Performance by Model.}
Model error rate (dashed), human error rate (solid), and z-scored human reaction time (right axis) plotted as a function of relative rotation angle, shown separately for each VLM. Shaded regions indicate 95\% confidence intervals.}
    \label{fig:rotation_individualmodels}
\end{figure}

\section{Appendix: Human Behavioral Experiments}
\label{app:human_behavioral_exp}

%TLG: have the labels correspond to the sections in the text

Human behavioral data for the geometric complexity, numerical estimation, and mental rotation tasks were collected via Prolific under an IRB-approved protocol. To maintain double-blind review standards, we do not include the consent form. In the geometric complexity task, participants selected the oddball from an array of geometric objects. In the numerical estimation task, they counted the total number of objects presented. In the rotation task, they determined whether a pair of objects were mirrored versions of each other. Participants were compensated at an approximate rate of $\$12$ per hour.

\subsection{Geoclidean}

\begin{figure}[h]
  \begin{center}
  \includegraphics[width=\linewidth]{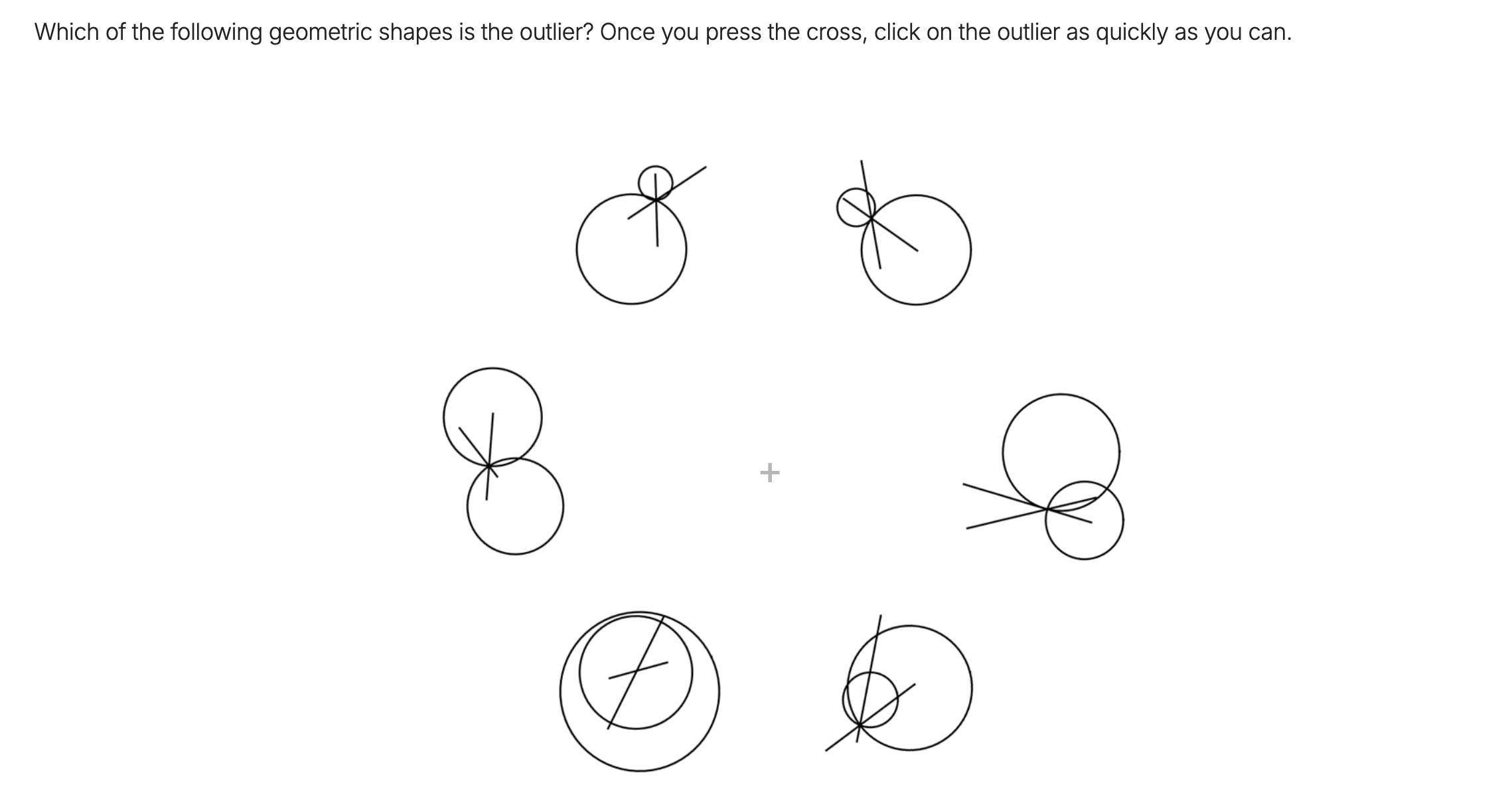}
  \end{center}
  \caption{\textbf{Example of Geoclidean Task.}}
    \label{fig:geo_example}
\end{figure}

\begin{figure}[h]
  \begin{center}
  \includegraphics[width=\linewidth]{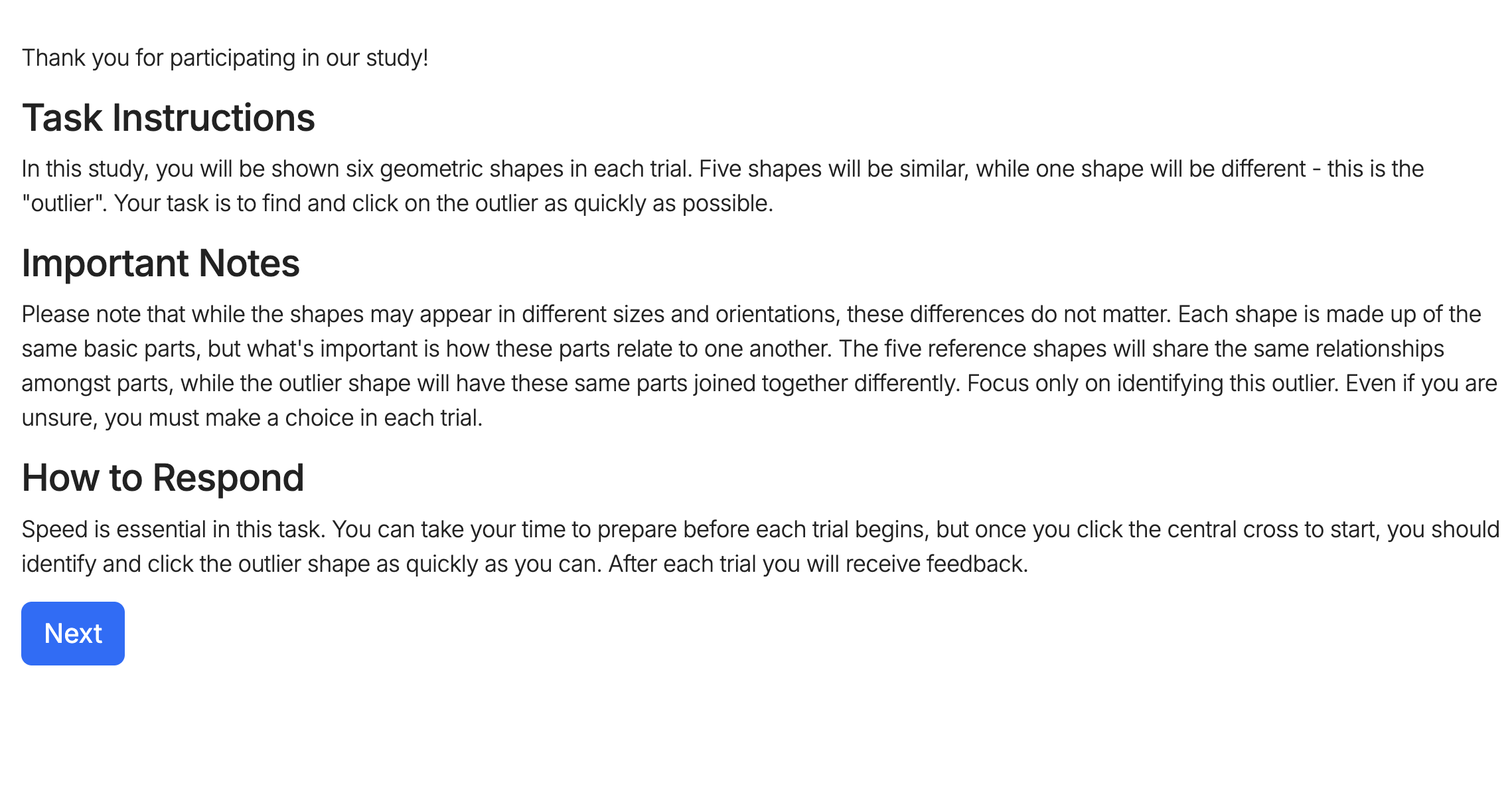}
  \end{center}
  \caption{\textbf{Instructions for Geoclidean Task.}}
    \label{fig:geo_instruction}
\end{figure}

\newpage

\subsection{Numerosity}

\begin{figure}[h]
  \begin{center}
  \includegraphics[width=\linewidth]{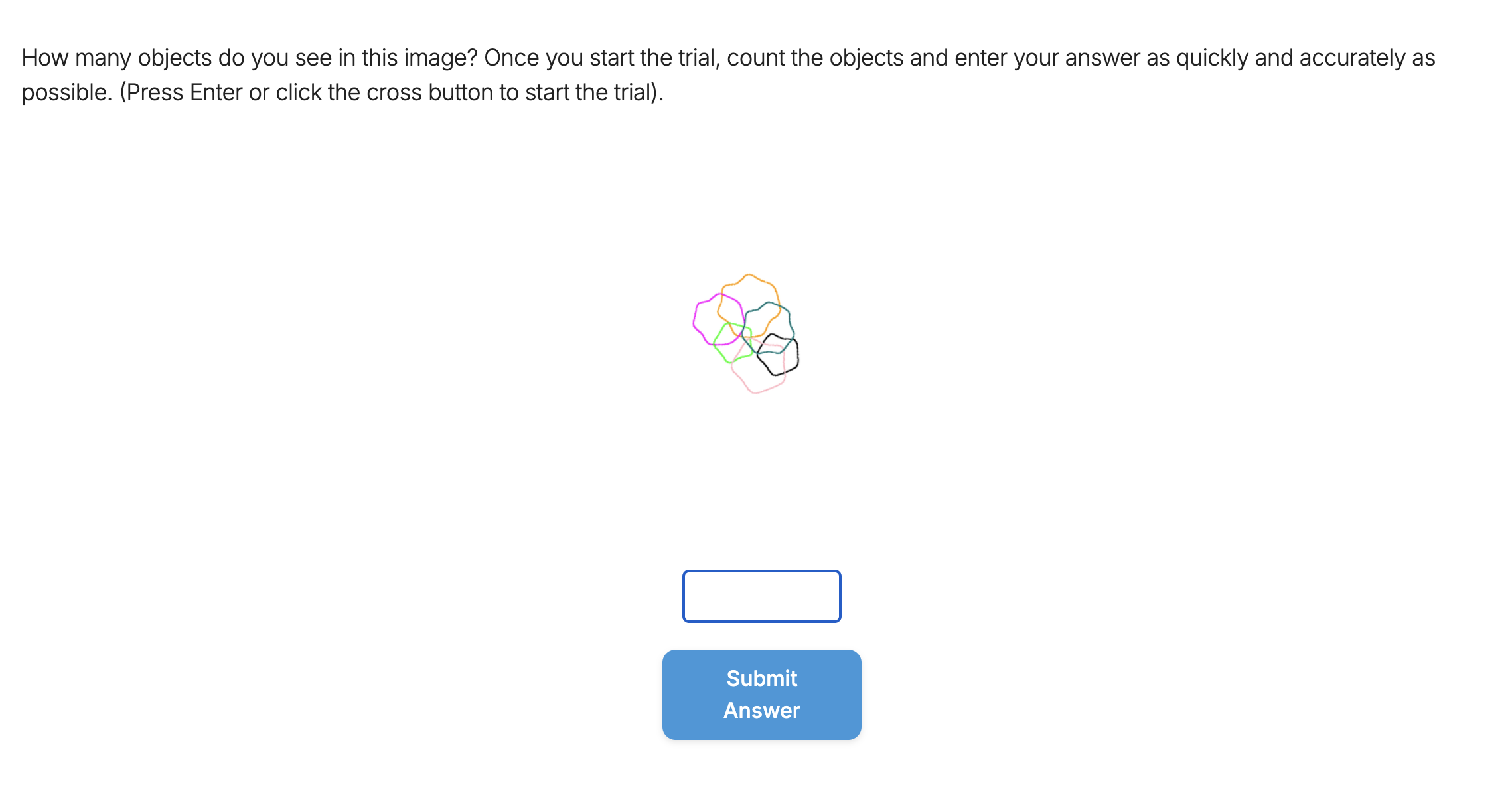}
  \end{center}
  \caption{\textbf{Example of Numerosity Task.}}
    \label{fig:num_example}
\end{figure}

\begin{figure}[h]
  \begin{center}
  \includegraphics[width=\linewidth]{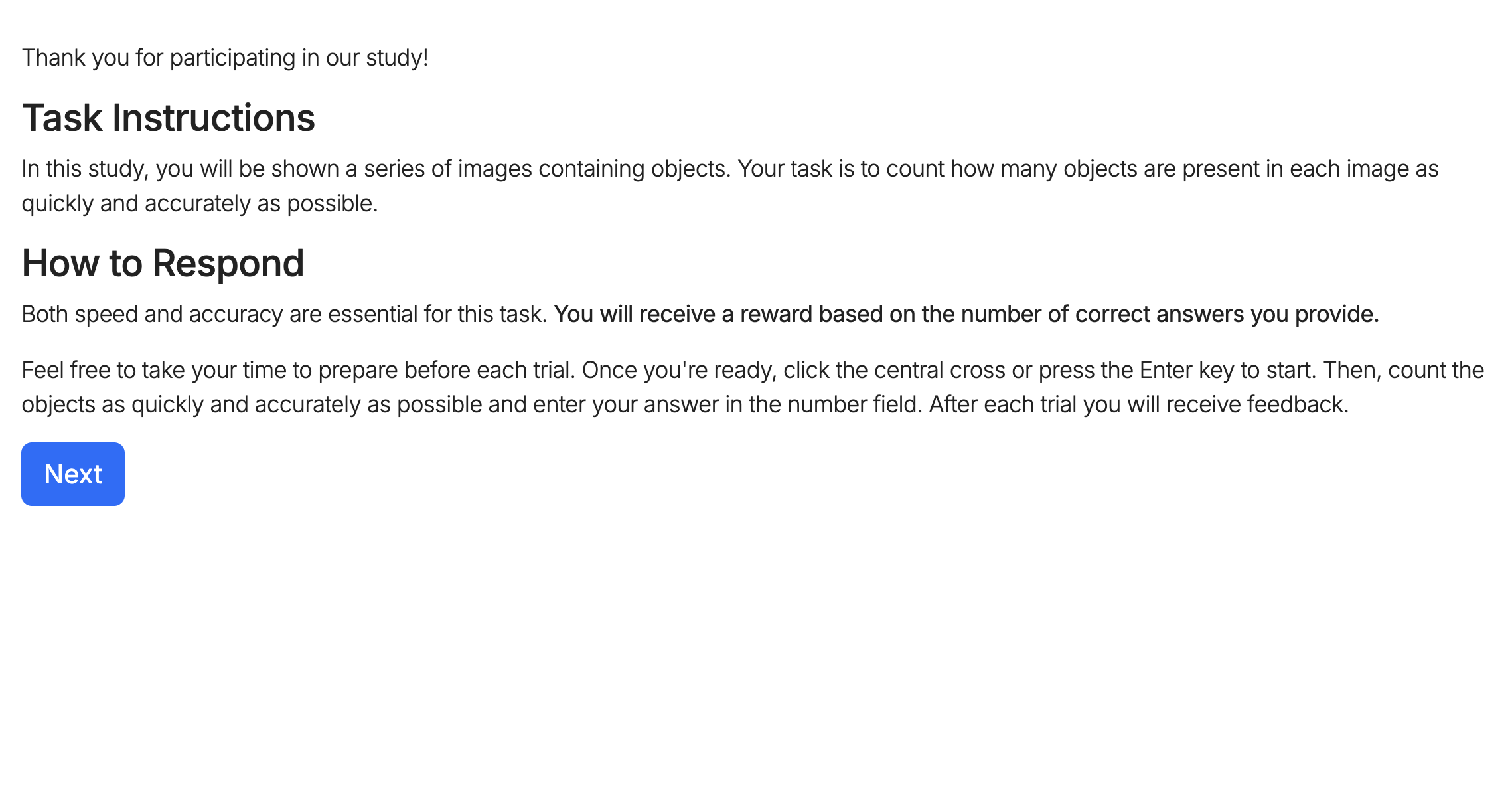}
  \end{center}
  \caption{\textbf{Instructions for Numerosity Task.}}
    \label{fig:num_instruction}
\end{figure}

\newpage

\subsection{Rotation}

\begin{figure}[h]
  \begin{center}
  \includegraphics[width=\linewidth]{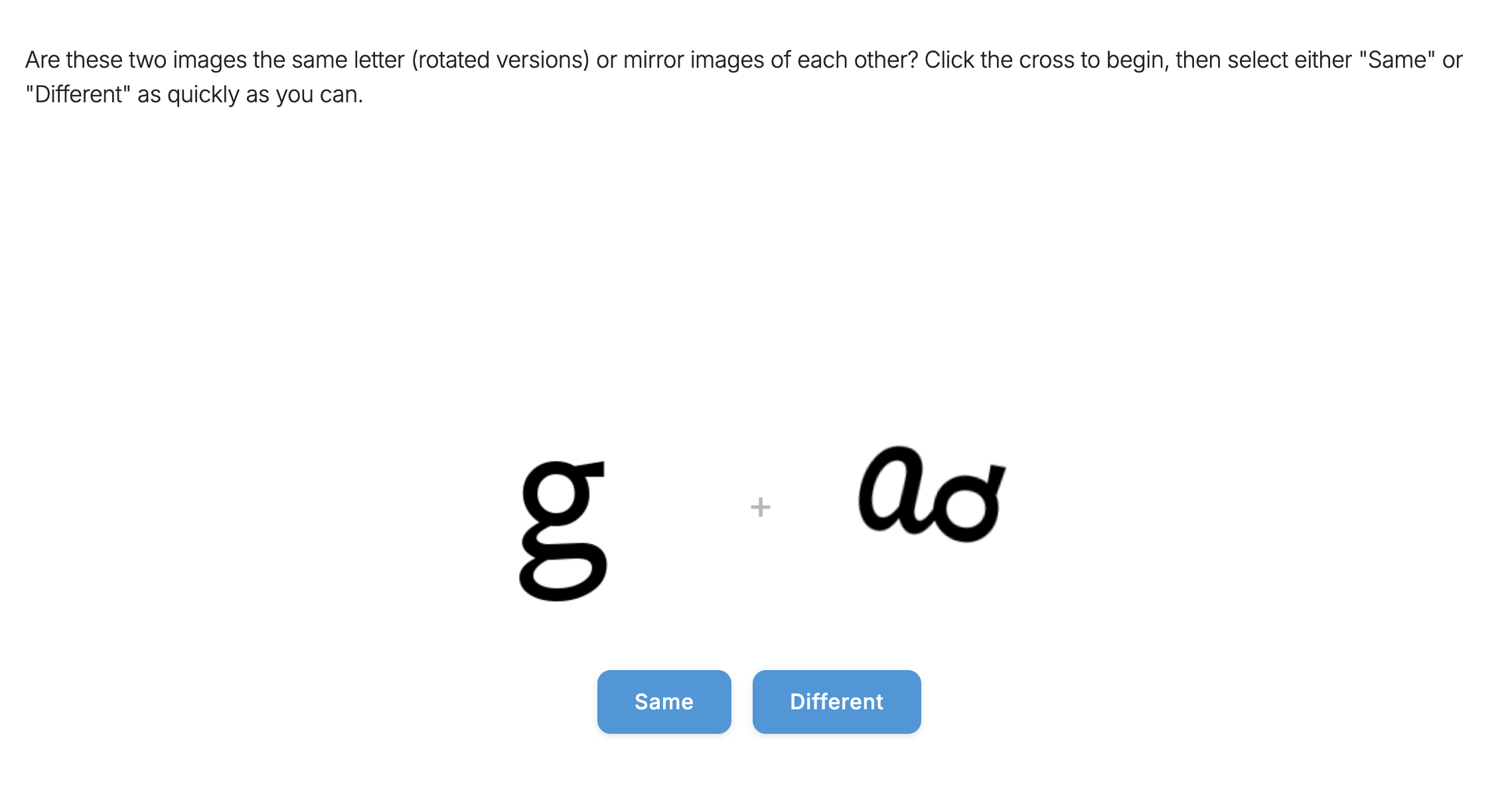}
  \end{center}
  \caption{\textbf{Example of Rotation Task.}}
    \label{fig:rotation_example}
\end{figure}

\begin{figure}[h]
  \begin{center}
  \includegraphics[width=\linewidth]{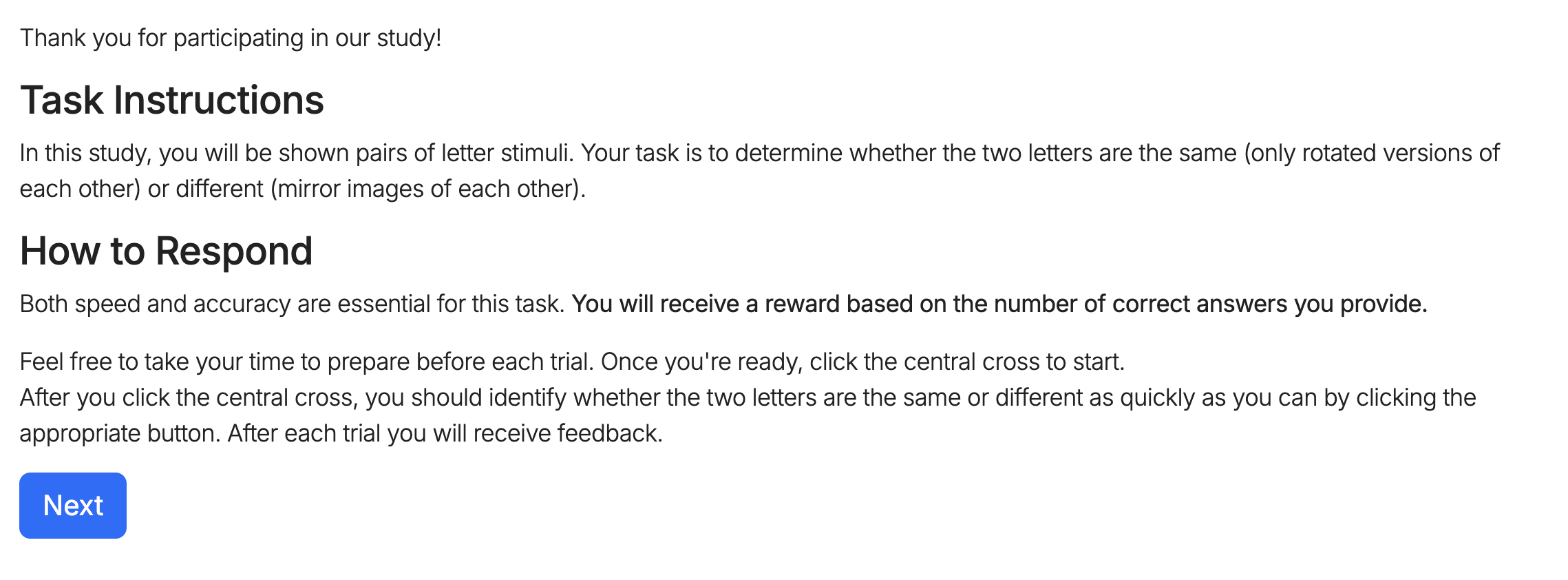}
  \end{center}
  \caption{\textbf{Instructions for Rotation Task.}}
    \label{fig:rotation_instruction}
\end{figure}

\newpage

\section{Appendix: Causal evidence for a serial processing deficit}

\subsection{Chain-of-Thought and Reasoning}
Chain-of-Thought experiments were conducted with GPT-4o, Claude Sonnet~3.7, Llama~4 Maverick, and Qwen2.5~VL, each prompted with the Chain-of-Thought templates described in Appendix~\ref{app:prompts}. Reasoning experiments were carried out with GPT-o3, Gemini~2.5~Pro, and Claude Opus~4, also using the Chain-of-Thought prompts. All three experimental paradigms were completed in full under the conditions specified in the main paper. The full set of results for these models, compared against humans, are reported in Table~\ref{table:expanded-results}.

\begin{table}[ht]
    \centering
        \renewcommand{\arraystretch}{1.5}
    \setlength{\tabcolsep}{10pt}
    \resizebox{\textwidth}{!}{%
    \begin{tabular}{|l|l|l|l|l|l|l|}
    \hline
        \textbf{Model} & Geoclidean (CI) & Numerosity (CI) & - & - & - & Rotation (CI) \\ \hline
        \textbf{} & ~ & uniform\_distinct & uniform\_overlapping & colored\_distinct & colored\_overlapping & ~ \\ \hline
        \textbf{Human} & 78.2 (77.5–78.9)\% & 98.7 (98.1–99.1)\% & 69.9 (68.3–71.5)\% & 98.5 (97.9–98.9)\% & 71.8 (70.2–73.4)\% & 79.6 (78.9–80.3)\% \\ \hline
        \textbf{NoCoT} & 66.9 (66.2–67.7)\% & 85.0 (83.8–86.2)\% & 43.9 (42.2–45.6)\% & 91.3 (90.3–92.2)\% & 61.7 (60.0–63.4)\% & 56.7 (55.9–57.5)\% \\ \hline
        \textbf{CoT} & 68.5 (67.7–69.2)\% & 89.2 (88.1–90.3)\% & 41.3 (39.6–43.0)\% & 96.0 (95.3–96.6)\% & 61.1 (59.4–62.7)\% & 56.9 (56.1–57.8)\% \\ \hline
        \textbf{Reasoning} & 76.2 (75.4–77.0)\% & 95.6 (94.7–96.3)\% & 43.9 (41.9–45.9)\% & 99.7 (99.5–99.9)\% & 72.1 (70.3–73.9)\% & 58.1 (57.1–59.0)\% \\ \hline
    \end{tabular}
    }
    \caption{Mean accuracy (95\% CI) for humans and all model classes (NoCoT (Baseline), CoT, and Reasoning) across the full evaluation set, spanning Geoclidean, Numerosity (four conditions), and Rotation tasks.}
    \label{table:expanded-results}
\end{table}

\subsection{Tool-Use}
Tool-use experiments were conducted through the ChatGPT interface with the GPT-o3 model. Across three experiments, we performed a total of 400 trials, focusing on challenging task conditions: 100 trials on Geoclidean tasks (20 for each of five tasks, including two with MDL~3 and three with MDL~4), 100 trials on Rotation tasks (angles from $90^\circ$ to $270^\circ$ in $30^\circ$ increments), and 200 trials on Numerosity tasks ($n=8$ objects, 50 trials for each of the four conditions). The model was prompted with the Chain-of-Thought prompts, as detailed in Appendix \ref{app:prompts}. Results on this restricted set of 400 challenging trials are shown in Table~\ref{table:tool-use-results}.

\begin{table}[ht]
    \centering
    \renewcommand{\arraystretch}{1.5}
    \setlength{\tabcolsep}{10pt}
    \resizebox{\textwidth}{!}{%
    \begin{tabular}{|l|l|l|l|l|l|l|}
    \hline
        Model & Geoclidean (CI) & Numerosity (CI) & - & - & - & Rotation (CI) \\ \hline
        ~ & ~ & uniform\_distinct & uniform\_overlapping & colored\_distinct & colored\_overlapping & ~ \\ \hline
        \textbf{Human} & 58.0 (55.7–60.2)\% & 96.9\% (92.3–98.8\%) & 73.5\% (66.3–79.6\%) & 93.8\% (88.7–96.7\%) & 73.9\% (65.4–81.0\%) & 75.2 (71.5–78.6)\% \\ \hline
        \textbf{NoCoT} & 45.5 (40.7–50.4)\% & 58.5\% (51.6–65.1\%) & 9.5\% (6.2–14.4\%) & 75.0\% (68.6–80.5\%) & 14.5\% (10.3–20.0\%) & 47.8 (43.1–52.5)\% \\ \hline
        \textbf{CoT} & 43.5 (38.7–48.4)\% & 70.9\% (64.2–76.7\%) & 6.0\% (3.5–10.2\%) & 78.4\% (72.2–83.5\%) & 18.1\% (13.4–24.0\%) & 47.3 (42.6–52.2)\% \\ \hline
        \textbf{Reasoning} & 49.3 (43.7–55.0)\% & 79.2\% (72.0–84.9\%) & 8.7\% (5.1–14.3\%) & 99.3\% (96.3–99.9\%) & 29.3\% (22.6–37.1\%) & 48.4 (43.0–53.9)\% \\ \hline
        \textbf{Tool Use} & 66.0 (56.3–74.5)\% & 100.0\% (92.9–100.0\%) & 10.0\% (4.3–21.4\%) & 100.0\% (92.9–100.0\%) & 28.0\% (17.5–41.7\%) & 88.8 (81.0–93.6)\% \\ \hline
    \end{tabular}
    }
    \caption{Mean accuracy (95\% CI) for humans and all model classes (NoCoT (Baseline), CoT, Reasoning, and Tool Use) evaluated on the same 400 challenging trials spanning Geoclidean, Numerosity (four conditions), and Rotation tasks.}
    \label{table:tool-use-results}
\end{table}

\section{Appendix: Computational Resources}
All model queries were conducted via API (e.g., Together), including open-weight models such as Llama. No specialized hardware was required. Dataset generation and preprocessing were lightweight and completed on standard CPU machines.

\end{document}